\documentclass[conference]{IEEEtran}
\IEEEoverridecommandlockouts
\usepackage{cite}
\usepackage{algorithmic}
\usepackage{amsmath,amssymb,amsfonts}
\usepackage{graphicx}
\usepackage{hyperref}
\usepackage{textcomp}
\usepackage{xcolor}
\def\BibTeX{{\rm B\kern-.05em{\sc i\kern-.025em b}\kern-.08em
    T\kern-.1667em\lower.7ex\hbox{E}\kern-.125emX}}

\begin{document}
\title{DeMansia: Mamba Never Forgets Any Tokens}

\author{\IEEEauthorblockN{Ricky Fang}
	\IEEEauthorblockA{Department of Mathematics \\
		Simon Fraser University\\
		Burnaby, Canada \\
	\href{mailto:tfa24@sfu.ca}{tfa24@sfu.ca}}
}

\maketitle

\begin{abstract}
This paper examines the mathematical foundations of transformer architectures, highlighting their limitations particularly in handling long sequences. We explore prerequisite models such as Mamba, Vision Mamba (ViM), and LV-ViT that pave the way for our proposed architecture, DeMansia. DeMansia integrates state space models with token labeling techniques to enhance performance in image classification tasks, efficiently addressing the computational challenges posed by traditional transformers. The architecture, benchmark, and comparisons with contemporary models demonstrate DeMansia's effectiveness. The implementation of this paper is available on GitHub at \href{https://github.com/catalpaaa/DeMansia}{https://github.com/catalpaaa/DeMansia}.
\end{abstract}

\section{Introduction}
The landscape of deep learning has been profoundly reshaped by the advent of transformer architectures \cite{attention}, which have established new benchmarks across a broad spectrum of applications, particularly in natural language processing and computer vision. Transformers leverage the mechanism of self-attention \cite{attention} to dynamically weigh the significance of different parts of input data, facilitating more nuanced and contextually aware interpretations than previous sequence-based models could achieve.

Despite their success, transformers \cite{attention} are not without their limitations. Chief among these is the computational intensity of the self-attention mechanism \cite{attention}, which scales quadratically with the length of the input sequence. This characteristic makes traditional transformers less suited to tasks involving very large input sizes or requiring real-time processing on hardware with limited capabilities. Recent innovations \cite{linformer, linear_attention, sparse_attention} have sought to address these challenges by modifying the attention mechanism to reduce computational overhead, but these approaches often involve trade-offs in terms of accuracy and model complexity \cite{attention_running_time}.

In this context, we introduce the DeMansia model, a novel architecture that integrates the benefits of the Mamba \cite{mamba} and Vision Mamba \cite{vim}, while also incorporating advancements in training pipeline from LV-ViT \cite{lv-vit} to enhance performance in image classification tasks. DeMansia is designed to tackle the delivery of high performance on resource-constrained environments. The architecture combined the concept of positional aware state space models with an innovative application of token labeling \cite{lv-vit} that maintain computational efficiency without compromising contextual richness of the model's understanding.

This paper details the development of DeMansia and evaluates its performance against established benchmark in the field. We provide a comprehensive comparison with existing models, demonstrating the effectiveness of DeMansia in image classification tasks and its potential as a promising solution for a wide range of applications in computer vision.

\section{Background}
The introduction of Transformer architectures, spearheaded by the groundbreaking work of \cite{attention}, has led to a flurry of state-of-the-art models across a variety of domains in natural language processing and beyond. At the heart of the Transformer's prowess lies its attention mechanism, which enables the model to focus on different parts of the input data for a given task, thus capturing intricate dependencies. However, despite its success, the computational complexity associated with the attention mechanism raises challenges, particularly when dealing with large input sequences.

\subsection{Single-Headed Attention}
The attention mechanism as described in \cite{attention} operates as follows. Given an input sequence $X \in \mathbb{R}^{n \times d_{model}}$, where $n$ is the sequence length and $d_{model}$ denotes both the dimension of embedding vectors and the internal dimensions of the Feed-Forward Networks. The algorithm first projects $X$ into query ($Q $), key ($K$), and value ($V$) matrices through learnable weight matrices $W^Q \in \mathbb{R}^{d_{model} \times d_Q}$, $W^K \in \mathbb{R}^{d_{model} \times d_K}$, and $W^V \in \mathbb{R}^{d_{model} \times d_V}$ and learnable bias vectors $b_Q, b_K, b_V \in \mathbb{R}^{d_{model}}$:

\begin{equation}
	\begin{split}
		Q & = W_QX + Broadcast_{d_Q}(b_q)^\top \\
		K & = W_KX + Broadcast_{d_K}(b_k)^\top \\
		V & = W_VX + Broadcast_{d_V}(b_v)^\top
	\end{split}
	\label{att eq 1}
\end{equation}
\cite{attention} \cite{attention_math}

We assume that $d_Q = d_K$.

The Scaled Dot-Product Attention \cite{attention} is then computed as follows:

\begin{equation}
	\begin{split}
		Attention(Q,K,V) & = softmax(\frac{QK^\top}{\sqrt{d_{model}}})V
	\end{split}
	\label{att eq 2}
\end{equation}
\cite{attention}

The first step in computing the attention is to calculate the dot product of the query matrix $Q$ with the transpose of the key matrix $K$. Given that both $Q$ and $K$ are derived from the same input $X$ and have dimensions $n \times d_K$, where $n$ is the sequence length and $d_K$ is the dimension of the key vectors, the resulting matrix $QK^\top$ is of dimension $n \times n$. Each element of $QK^\top$ represents the attention score between a pair of query and key vectors, calculated using $2d_K$ operations (multiplications and additions).

Following the attention score computation, each element of $QK^\top$ is scaled by dividing by $\sqrt{d_{model}}$. This scaling is crucial for maintaining numerical stability during the $softmax()$ calculation, particularly because the dot product can grow large with increasing dimensions of $d_{model}$. The division is applied element-wise across the $n \times n$ matrix.

The $softmax()$ function is then applied row-wise across the scaled scores matrix $\frac{QK^\top}{\sqrt{d_{model}}}$. This operation converts the raw scores into a matrix of attention probabilities, indicating how much each output element should attend to every other element in the sequence. The $softmax()$ operation effectively normalizes the scores so that they sum to one across each row.

Finally, the matrix of attention probabilities is used to compute a weighted sum of the value vectors. This step involves multiplying the attention probabilities matrix by the value matrix $V$ of dimensions $n \times d_V$. The result is a matrix of the same dimensions, where each row is a weighted combination of all value vectors, tailored to the specific attention needs of each output element in the sequence.

This step directly constructs each output element from a context-aware, dynamically weighted combination of input features. The mechanism allows the Transformer \cite{attention} to selectively focus on different parts of the input sequence, extracting and emphasizing information that is most relevant for each part of the output. 

The above algorithm is also known as the single-headed attention.

The computation of the dot-product $QK^\top$ involves $n^2$ elements, each computed using $2d_K$ operations. Hence, the complexity for this step is $\mathcal{O}(n^2d_K)$. Each element of the $n \times n$ matrix is scaled element-wise, adding a complexity of $\mathcal{O}(n^2)$. The $softmax()$ function, applied row-wise, processes $n$ vectors of length $n$, adding an additional complexity of $\mathcal{O}(n^2)$. The final multiplication of the attention probabilities with the value matrix $V$ involves element-wise multiplications for each of the $n^2$ entries with a dimension of $d_V$, culminating in a complexity of $\mathcal{O}(n^2d_V)$. The overall time complexity of the attention mechanism can be expressed as $\mathcal{O}(n^2(d_K + d_V)) \in \mathcal{O}(n^2)$

\subsection{Multi-Head Attention}
\cite{attention} has also proposed multi-head attention with the number of head $h$. The input $X$ is linear projected with discrete, learnable weights matrices and bias vectors for each of the head:

\begin{equation}
	\begin{split}
		Q^{[h]} & = W_Q^{[h]}X + Broadcast_{d_Q}({b_q^{[h]}}^\top) \\
		K^{[h]} & = W_K^{[h]}X + Broadcast_{d_K}({b_k^{[h]}}^\top) \\
		V^{[h]} & = W_V^{[h]}X + Broadcast_{d_V}({b_v^{[h]}}^\top)
	\end{split}
	\label{att eq 3}
\end{equation}
\cite{attention} \cite{attention_math}

Each triple of $Q^{[h]}$, $K^{[h]}$ and $V^{[h]}$ is then have their attention computed using \eqref{att eq 2}, we denote the attention output of each head $Y^{[h]}$. For the multi-head attention, we combine the attention of each head using a learnable weight matrix $W_O \in \mathbb{R}^{hd_V \times d_{model}}$ and a learnable bias vector $b_O \in \mathbb{R}^{d_v}$:

\begin{equation}
	\begin{gathered}
		Y = [Y^{[1]}, Y^{[2]}, \dots, Y^{[h]}] \\
		MultiHead(Q,K,V) = W_OY + Broadcast_{hd_V}(b_O^\top)
	\end{gathered}
	\label{att eq 4}
\end{equation}
\cite{attention} \cite{attention_math}

This concatenation step combines the independently attended features from each head into a single matrix, effectively mixing the different learned representations from each head.

In the multi-head attention \cite{attention}, the single-headed attention \cite{attention} is ran for $h$ heads, leading to a runtime of $\mathcal{O}(hn^2(d_K+d_V)) \in \mathcal{O}(n^2)$.

\begin{figure}[!htbp]
	\centering
	\includegraphics[width=0.8\linewidth]{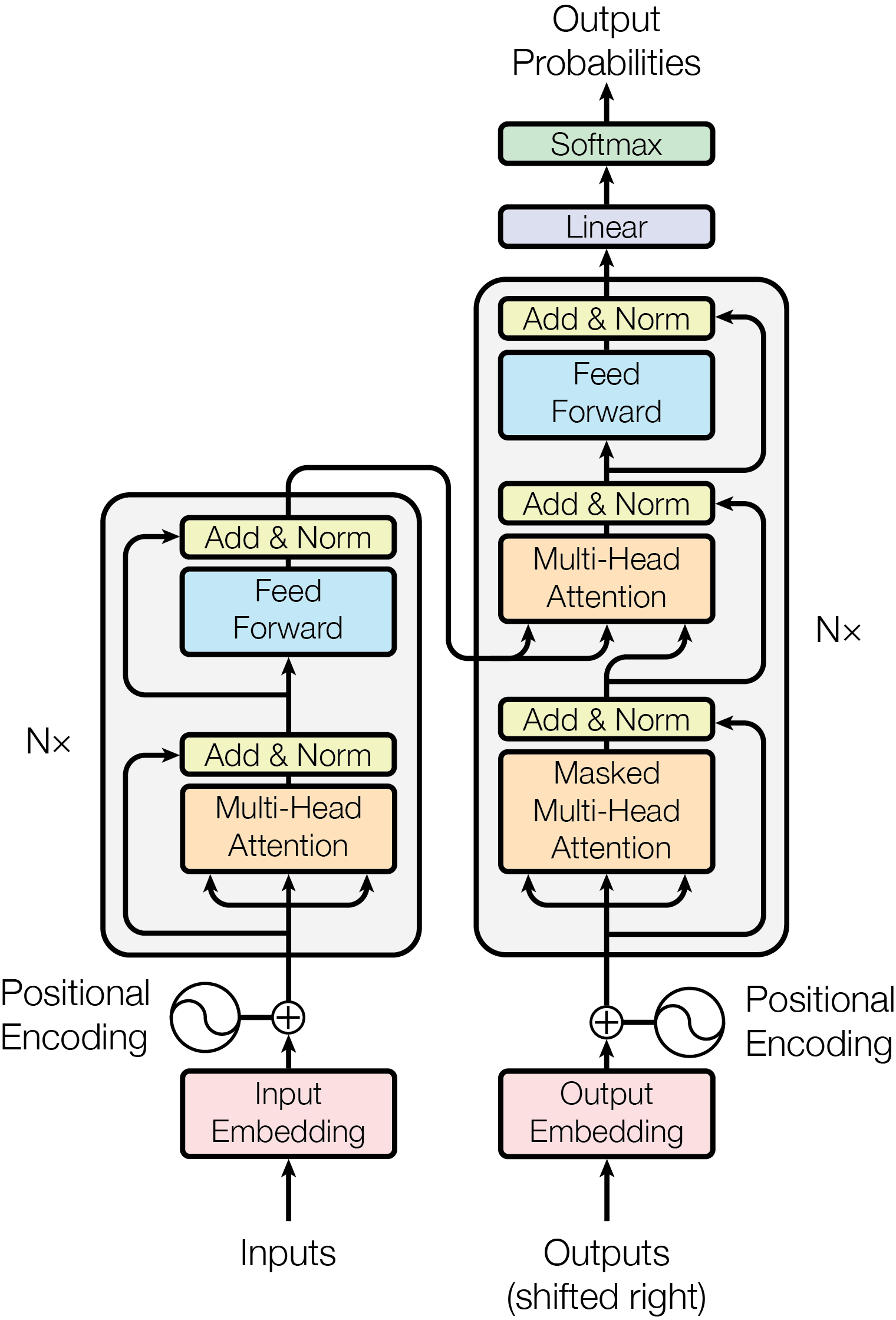}
	\caption{Transformer architecture proposed by \cite[Figure 1]{attention}.}
	\label{fig:transformer pipeline}
\end{figure}

The standard transformer block \cite{attention} utilize the multi-head attention \cite{attention} as shown in Fig. \ref{fig:transformer pipeline}. The quadratic scaling with respect to the input length poses significant computational bottlenecks when applying transformer to fields such as computer vision, where inputs can reach billions of pixels. To manage computational demands, models are often constrained in terms of parameter size or have input compression, limiting their performance.

Although recent studies propose optimizations to achieve linear or near linear time complexity in the attention mechanism. For instance, \cite{linformer} reduces the complexity of the attention mechanism to $\mathcal{O}(n)$ by using low-rank matrix approximations. \cite{linear_attention} addresses the computational intensity by transforming the attention computation into a linear dot-product of kernel feature maps with linear runtime, and Sparse Transformers \cite{sparse_attention} employ sparse factorizations to lower the time complexity of the attention mechanism to $\mathcal{O}(n \sqrt{n})$. These adaptations introduce approximation errors compared to the original formulation \cite{attention_running_time}.

Given the computational limitations imposed by the attention mechanism \cite{attention}, especially for achieving state-of-the-art performance on high-resolution inputs or on less powerful consumer-grade devices, there is a need for more efficient approaches beyond the canonical Transformer model.

\section{Related work}
Our work build on prior work in several domains: state space model (SSM) architecture Mamba \cite{mamba}, Vision Mamba (ViM)'s \cite{vim} bi-directional Mamba block and token labeling training pipeline from VL-ViT \cite{lv-vit}.

\subsection{Mamba}
To appreciate the novelty of  Mamba \cite{mamba}, it is essential to first understand the structured state space sequence models (S4) \cite{s4} and selective scan S4 (S6) \cite{mamba}.

The first stage of the S4 \cite{s4} model is governed by a combination of linear transformations represented by the equations:

\begin{equation}
	\begin{split}
		h'(t) & = Ah(t) + Bx(t) \\
		\quad y(t) & = Ch(t)
	\end{split}
	\label{s4 eq 1}
\end{equation}
\cite{s4} \cite{mamba}

where $h(t)$ represents the hidden state at time $t$, $x(t)$ denotes the input, and $y(t)$ is the output. The parameters $A$, $B$, and $C$ facilitate the transformation of inputs into a latent representation and subsequently to outputs. Note that $A$, $B$, and $C$ are shared parameters across all hidden states.

The second stage involves discretizing these continuous parameters ($A$, $B$, $C$) into their discrete counterparts ($\bar{A}$, $\bar{B}$, $\bar{C}$) with respect to a timescale parameter $\Delta$.

\begin{equation}
	\begin{split}
		\bar{A} & = exp(\Delta A) \\
		\bar{B} & = (\Delta A)^{-1}(exp(\Delta A) - I) \cdot \Delta B
	\end{split}
	\label{s4 eq 2}
\end{equation}
\cite{s4} \cite{mamba}

Discretization allows the model to operate within discrete time steps effectively. This transforms \eqref{s4 eq 1} into:

\begin{equation}
	\begin{split}
		h_t & =\bar{A}h_{t-1} + \bar{B}x_t \\
		y_t & = Ch(t)
	\end{split}
	\label{s4 eq 3}
\end{equation}
\cite{s4} \cite{mamba}

Finally, the recurrent relation $h_t$ can be written as a convolution:

\begin{equation}
	\begin{split}
		\bar{K} & = (C\bar{B}, C\bar{A}\bar{B}, \dots , C\bar{A}^{M-1}\bar{B}) \\
		y & = x * \bar{K}
	\end{split}
	\label{s4 eq 4}
\end{equation}
\cite{s4} \cite{mamba}

where $M$ is the length of the input sequence $x$.

However, S4 \cite{s4} compresses it's context into a smaller state, while efficient, poses limitations when dealing with extremely long sequences or requiring more complex, dynamic interactions within input \cite{mamba}. S4 is also time and input invariant, leading to low content aware reasoning performance \cite{mamba}.

The subsequent development, S6 \cite{mamba}, sought to address these limitations by introducing selectivity in the state space model to maintain the uncompressed context. S6 follows \eqref{s4 eq 1}, but define matrices $B, C$ as follow.

\begin{equation}
	\begin{split}
		B & = linear_N(x) \\
		C & = linear_N(x)
	\end{split}
	\label{s6 eq 1}
\end{equation}
\cite{s4} \cite{mamba}

where $linear_d$ is a learnable parameterized projection to dimension $d$

The discretization of \eqref{s6 eq 1} follows \eqref{s4 eq 2}, but with a change to $\Delta$ as follow:

\begin{equation}
	\begin{split}
		\Delta & = softplus(Param. + Broadcast_{d_{model}}(Linear_1(x)))
	\end{split}
	\label{s6 eq 2}
\end{equation}
\cite{s4} \cite{mamba}

Since $B$, $C$ and $\Delta$ are relative to the input, not the hidden state of the previous layer, the computation for $B$, $C$ and $\Delta$ can be calculated efficiently in parallel. The transformation matrix $A$ for the hidden states remains unchanged and constant for each layer.

The selection mechanism \cite{mamba} functions analogously to attention mechanisms. Like attention, which dynamically allocates computational focus to different segments of the input based on their relevance to the task, the selection mechanism selectively utilizes components of the input to construct the high dimension state space. Comparing to to S4 \cite{s4}, the selection mechanism allows the model to be time and context aware.

Mamba \cite{mamba} improves upon S6 \cite{mamba} by substituting the convolution operation in \eqref{s4 eq 3} with a scan operation, also known as a prefix sum \cite{prefix_sum} operation. This modification introduces a more parallelizable approach, streamlining computation and enabling faster processing of sequences.

Mamba goes one step further by incorporating hardware-aware optimizations. It schedules the GPU's High Bandwidth Memory (HBM) access during the scan operation to mitigate unnecessary data transfers between the HBM and the GPU's Static Random-Access Memory. With these optimizations, Mamba can reach linear scaling with high performance on long context length, the detail of which can be found in \cite[Appendix D]{mamba}.

\begin{figure}[!htbp]
	\centering
	\includegraphics[trim={0 0 3.5cm 0}, clip, width=0.5\linewidth]{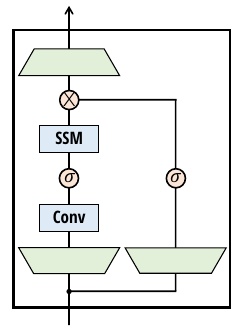}
	\caption{Mamba block proposed by \cite[Figure 3]{mamba}. The Trapezoid is a linear projection layer. \textcircled{$\sigma$} is SiLU / Swish activation \cite{GELU, swish}. \textcircled{$\times$} is multiplicative gate.}
	\label{fig:mamba block pipeline}
\end{figure}

\subsection{ViM}
While the original Mamba architecture has shown considerable promise in the domain of language modeling \cite[Table 3]{mamba}, its extension to spatially-aware tasks such as computer vision has necessitated further innovation. \cite{vim} argues that to effectively process image data, recognizing spatial relationships is critical—a capability that the original Mamba block \cite{mamba} does not inherently possess. To address this limitation, the ViM block \cite{vim} was proposed as a novel adaptation.

\begin{figure}[!htbp]
	\centering
	\includegraphics[width=\linewidth]{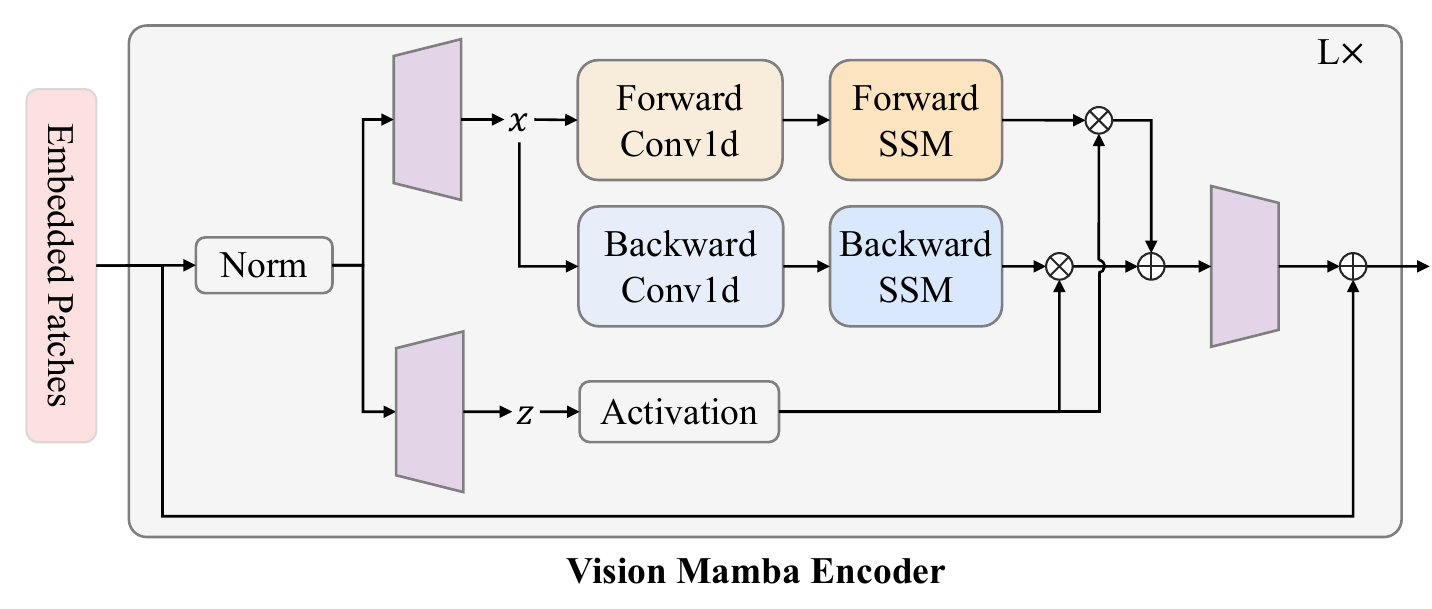}
	\caption{ViM block proposed by \cite[Figure 2]{vim}. The Trapezoid is a linear projection layer.}
	\label{fig:vim block pipeline}
\end{figure}

The enhancement introduced by the ViM block \cite{vim} lies in its bidirectional SSM mechanism. After the first linear projection, the input sequence undergoes two parallel processing by 1-D convolution and the SSM, in the forward and backward directions respectively. The results from these two passes are then merged. \cite{vim} demonstrates that approach enables the model to encompass and leverage spatial correlations within the data.

\begin{figure}[!htbp]
	\centering
	\includegraphics[width=0.9\linewidth]{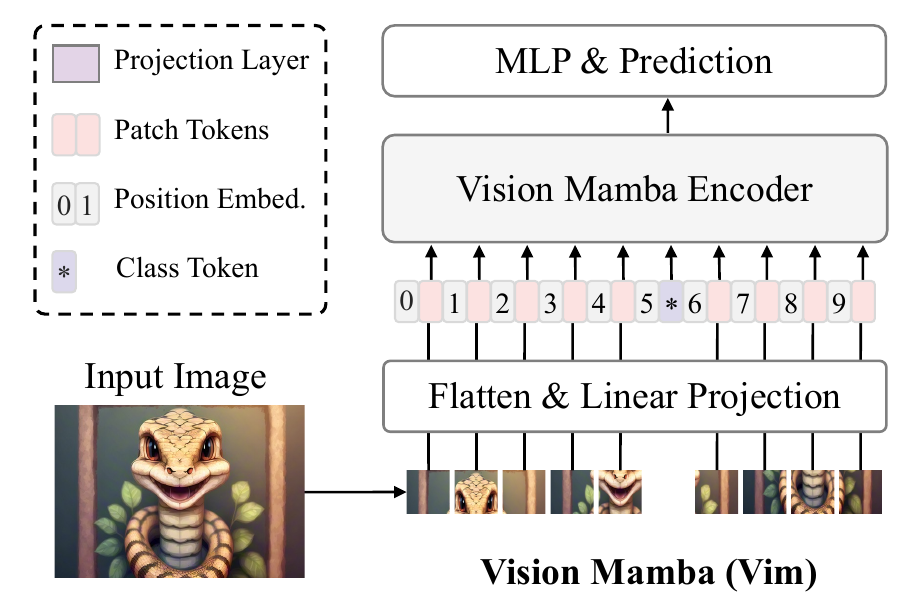}
	\caption{ViM architecture proposed by \cite[Figure 2]{vim}.}
	\label{fig:vim pipeline}
\end{figure}

For image classification tasks, the ViM \cite{vim} architecture follows that of the Vision Transformer \cite{vit}, with the primary modification being the replacement of the standard transformer block \cite{attention_math} with a ViM block, as illustrated in Fig. \ref{fig:vim pipeline}.

Given a input image $t \in \mathbb{R}^{H \times W \times C}$ where $H$, $W$, and $C$ denote the height, width, and channel count of the image, respectively. We first flatten $t$ into $J$ two dimensional patches $x_P \in \mathbb{R}^{j \times (p^2 \cdot C)}$ where P is the size of each image patches in pixels. Each $x_P$ is then linearly projected to vector with size $d_model$ by finding $t^j_pW$ where $t^j_p$ is the $j$-th patch of $t$ and $W$ is a learnable projection matrix. We concatenate all $t^j_p$ as well as a class embedding $t_{cls}$ of the same shape into a matrix $\in \mathbb{R}^{(J + 1) \times d_{model}}$, then we add a learnable positional embedding $E_{pos}$ of the same shape to the matrix, that gives our first embedding sequence

\begin{equation}
	T_0 = [t_{cls}; t^1_pW; t^2_pW; \dots; t^J_pW] + E_{pos}
	\label{vim eq 1}
\end{equation}
\cite{vim}

This sequence feeds into the first ViM block \cite{vim} to compute $T_1$, and the subsequent $T_{l-1}$ embedding sequences are feed to the $l$-th ViM block, until we reach the final ViM block. The class embedding is isolated and normalized post the final ViM block, and subsequently input into a Multi-Layer Perceptron head with input dimension $d_{model}$ that maps to the class output dimensions.

It is noteworthy that the placement of $t_{cls}$ within the sequence can vary, and introducing multiple class embeddings could also influence the model's performance. For detailed analyses on these configurations and their impacts on ViM's \cite{vim} accuracy, readers are directed to consult \cite{vim}.

In \cite{vim}, when comparing ViM-tiny \cite{vim} and DeiT-Ti \cite{deit}, two models with similar parameter count, the authors noted a $73.2\%$ reduction in VRAM usage and $2.8\times$ speedup during inferencing on image of size $1248^2$.

\subsection{LV-ViT}
In the original ViT \cite{vit}, and by extension ViM \cite{vim}, the final class prediction and corresponding loss are derived from the class token. We denote the output tokens of the final transformer block in ViT as $[x^{cls}, x^1, x^2, \dots, x^j]$. LV-ViT \cite{lv-vit} utilize the patch tokens to further improve the models accuracy on image classification.

Given a image $t$, we first prepare a dense map of the ground truth class on $t$. This map is partitioned into patches analogous to those in ViT \cite{vit}, represented as $[y^1, y^2, \dots, y^J]$. These dense score patches serve as ground truth for their corresponding ViT patch tokens, $x^j$. The auxiliary loss, derived from the cross-entropy between each patch token and its corresponding ground truth, is formulated as:

\begin{equation}
	L_{tl} = \frac{1}{N}H(x^{j}, y^{j})
	\label{lv-vit eq 1}
\end{equation}
\cite{lv-vit}

where $H$ is the cross-entropy loss. This auxiliary loss is added with the global class token cross-entropy loss, weighted by a coefficient $\beta$, yielding the total loss equation:

\begin{equation}
	L_{total} = H(x^{cls}, y^{cls}) + \beta \times \frac{1}{N}H(x^{j}, y^{j})
	\label{lv-vit eq 2}
\end{equation}
\cite{lv-vit}

\begin{figure*}[!htbp]
	\centering
	\includegraphics[width=0.7\linewidth]{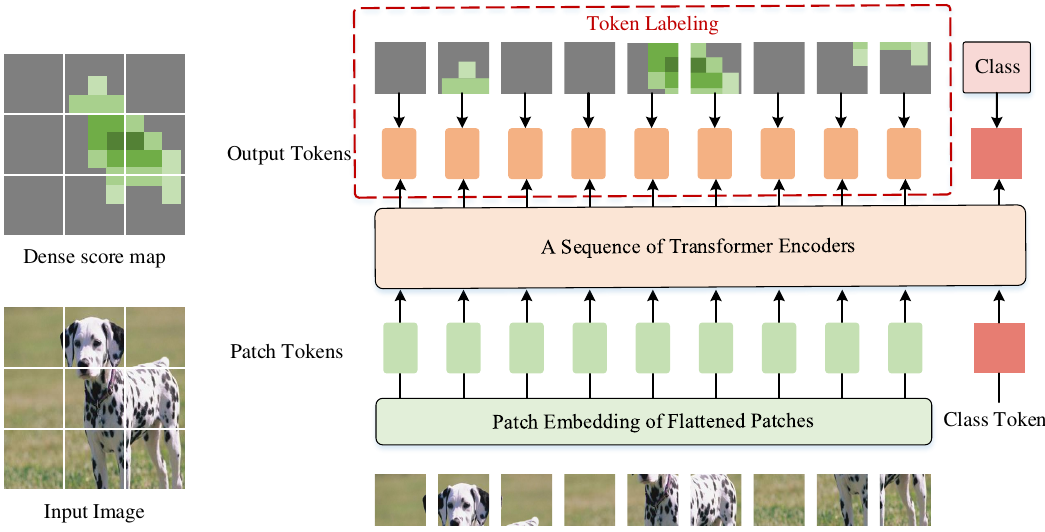}
	\caption{Pipeline of training vision transformers with token labeling. \cite[Figure 2]{lv-vit}}
	\label{fig:lv-vit training pipeline}
\end{figure*}

Here, $\beta$ acts as an adjustable hyper-parameter as a way to balance the two terms. This training loss pipeline is demonstrated visually in Fig. \ref{fig:lv-vit training pipeline}.

LV-ViT \cite{lv-vit} improves ViT \cite{vit} for several percentages on the validation accuracy on the , the detail comparison between models can be found in \cite[Table 4]{lv-vit}.

\section{The DeMansia model}
The DeMansia model architecture adheres closely to that of the ViM \cite{vim}. DeMansia initiating its process with a four-layer convolutional network that transforms an input image into a sequence of patch embeddings. Each embedding in this sequence is added with a learnable positional embedding to retain spatial information. Additionally, a learnable class embedding is placed in the center of the sequence and augmented by its corresponding positional embedding.

Once the initial sequence is assembled, it is advanced through multiple ViM blocks \cite{vim}. After transiting through the last ViM block, the resulting output is separated into patch and class components. These components are then individually processed along two distinct feed-forward layers. We denote one as the Aux Head and the other as the Class Head. The patch embeddings are processed by the Aux Head, which uses the same learnable parameters for all patch embeddings to calculate patch tokens. The class embedding is directed through the Class Head, which has its discrete learnable parameter to for calculating global class token.

During the training phase, in alignment with the LV-ViT \cite{lv-vit} methodology, DeMansia computes the token labeling loss as detailed in Equation \eqref{lv-vit eq 2}. In the inference stage, the model calculates the global class token as:

\begin{equation} \label{demansia eq 1}
	pred = x^{cls} + 0.5 \times max({x^1, x^2, \dots, x^j})
\end{equation}

Here, $x^{cls}$ is the class token calculated by the Class head and the set ${x^1, x^2, \dots, x^j}$ is the set of patch tokens outputted by the Aux Head.

We proved the following models shown in Tabel \ref{tabel:DeMansia model variants}.

\begin{table*}[!htbp]
	\caption{DeMansia model variants.}
	\centering
	\begin{tabular}{|c|c|c|c|c|}
		\hline
		Name               & Model Dim & Number of Layers & Number of parameters & Input Resolution \\
		\hline
		DeMansia Tiny      & 192       & 24               & 8.06M                & $224^2$          \\
		DeMansia Tiny 384  & 192       & 24               & N/A                  & $384^2$          \\
		DeMansia Small     & 384       & 24               & N/A                  & $224^2$          \\
		DeMansia Small 384 & 384       & 24               & N/A                  & $384^2$          \\
		\hline
	\end{tabular}
	\label{tabel:DeMansia model variants}
\end{table*}

\section{Experiments}

\subsection{Experiments Setup}
Due to constraints in time and computational resources, our experiments were limited to the DeMansia Tiny variant. The focus was narrowed to challenge the model on image classification tasks exclusively.

We employed the ImageNet-1k dataset \cite{imagenet} for our experiments. This dataset comprises approximately 1.28 million training images and 50,000 validation images, distributed across 1,000 classes. Additionally, we utilized the dense class map dataset \cite{lv-vit} prepared using the NFNet-F6 \cite{nfnet}, which provides detailed class dense map for the entire ImageNet-1k training set. The datasets were shuffled at the start of each epoch to improve generalization of the model.

For each image in the ImageNet-1k dataset \cite{imagenet}, we applied random crop augmentations and random flip augmentations. The image is then normalized to size $224^2$.

DeMansia Tiny was trained for 310 epochs using the RAdam optimizer \cite{radam} with a weight decay of 0.05. The learning rate was initiated at $1 \times 10^{-3}$ and scheduled by Cosine Annealing with Warm Restarts \cite{SGDR}, setting $T_0 = 10$ and $T_{mult} = 2$. We also utilized exponential moving average updates for model parameters at the closure of each epoch.

The experiments were conducted using a single RTX A6000 GPU with a batch size of 768. We leveraged automatic mixed precision (AMP) \cite{mixed_precision}, operating in Bfloat16 \cite{bfloat16} format for computation while preserving model weights in Float32.

We provide code and pretrained models to reproduce our experiments at \href{https://github.com/catalpaaa/DeMansia}{https://github.com/catalpaaa/DeMansia}.

\subsection{Result}

\begin{table*}[!htbp]
	\caption{Comparison of different models with a similar scale on ImageNet-1K \cite{imagenet} validation set. We use \cite{pytorch_benchmark} to record number of parameters, GFLOPs and memory usage for each model.}
	\begin{center}
		\begin{tabular}{|c|c|c|c|c|c|c|c|}
			\hline
			Name                        & Num. of Parameters & Input Resolution & Top-1 Accuracy (\%) & Top-5 Accuracy (\%) & GFLOPs        & Memory Usage (kB) \\
			\hline
			\textbf{DeMansia Tiny}      & \textbf{8.06M}     & $\mathbf{224^2}$ & \textbf{79.4}         & \textbf{94.5}         & \textbf{1.24} & \textbf{8024.58}  \\
			LV-ViT-S \cite{lv-vit}      & 26.15M             & $224^2$          & 83.3                & N/A                 & 6.01          & 10866.18          \\
			TinyViT-21M \cite{tiny-vit} & 21.20M             & $224^2$          & 83.1                & 96.5                & 4.07          & 144028.67         \\
			ViT-S \cite{vit}            & 22.05M             & $224^2$          & 81.0                & N/A                 & 4.24          & 4886.53           \\
			TinyViT-11M \cite{tiny-vit} & 11.00M             & $224^2$          & 81.3                & 95.8                & 1.89          & 140439.55         \\
			ViM-small \cite{vim}        & 25.80M             & $224^2$          & 80.5                & 95.1                & 0.06          & 8125.44           \\
			DeiT-small \cite{deit}      & 22.05M             & $224^2$          & 79.9                & 95.0                & 4.24          & 4886.53           \\
			LV-ViT-T \cite{lv-vit}      & 8.53M              & $224^2$          & 79.1                & N/A                 & 2.61          & 8394.24           \\
			ResNet-152 \cite{resnet}    & 60.19M             & $224^2$          & 78.6                & 95.5                & 11.58         & 143036.42         \\
			ViM-tiny \cite{vim}         & 7.15M              & $224^2$          & 76.1                & 93.0                & 0.03          & 4241.92           \\
			TinyViT-5M \cite{tiny-vit}  & 5.39M              & $224^2$          & 75.3                & 94.8                & 1.18          & 140570.62         \\
			DeiT-tiny \cite{deit}       & 5.72M              & $224^2$          & 72.2                & 91.1                & 1.07          & 3013.63           \\
			\hline
		\end{tabular}
	\end{center}
	\label{tabel:model comparison}
\end{table*}

The experimental results, as summarized in Table \ref{tabel:model comparison}, present a detailed comparison of DeMansia Tiny with other models that are either ConvNet-based, Transformer-based, or SSM-based, and of a similar scale. This comparison provides valuable insights into the competing strengths of these architectures, especially in relation to DeMansia Tiny's performance.

DeMansia Tiny achieves a top-1 accuracy of 79.4\% and a top-5 accuracy of 94.5\%. This performance demonstrates DeMansia's ability to effectively handle image classification tasks, positioning it competitively among models of similar scale.

When compared to other Transformer-based models, DeMansia Tiny demonstrates strong performance, though it falls short against models with significantly larger parameter counts. For instance, ViT-S \cite{vit}, with 22.05 million parameters, achieves a top-1 accuracy of 81.0\%.

Similarly, TinyViT-21M \cite{tiny-vit}, containing 21.20 million parameters, achieves a top-1 accuracy of 83.1\%. The superior accuracy of TinyViT-21M is partly due to the knowledge distillation technique employed during its training process. LV-ViT-S \cite{lv-vit}, with its 26.15 million parameters, also surpass DeMansia with a top-1 accuracy of 83.3\%. Compared to these two models, DeMansia Tiny’s performance is commendable given its significantly lower parameter count and resource usage.

DeiT-small \cite{deit}, also with 22.05 million parameters, reaches a top-1 accuracy of 79.9\%. DeMansia Tiny’s near equivalent accuracy with fewer parameters showcases its effective use of resources.

TinyViT-11M \cite{tiny-vit}, with 11 million parameters, achieves a top-1 accuracy of 81.3\%. Like its larger counterpart, TinyViT-11M benefits from knowledge distillation. DeMansia Tiny requires future fine-tuning to approach similar performance.

Among SSM-based models, DeMansia Tiny stands out. It surpasses the ViM-tiny model and closely matches the performance of ViM-small, showcasing the advantages of the token labeling used in DeMansia. However, the added VRAM and computational overhead due to the aux head is notable.

Comparing to ConvNet-based architectures like ResNet-152, which has approximately 7.5 times more parameters than DeMansia Tiny, our model demonstrates competitive performance. Specifically, the top-1 accuracy of DeMansia Tiny surpasses ResNet-152, which is notable given the significant difference in the number of parameters and GFLOPs.

While DeMansia Tiny excels in certain areas, it is essential to acknowledge its limitations. The added computational and memory overhead due to the bidirectional nature of the ViM block makes it more resource-intensive than some other models in its parameter range. Additionally, further fine-tuning and optimization could enhance DeMansia's performance, potentially closing the gap with larger models.

Overall, while DeMansia Tiny outperform some of the larger Transformer-based models and excels at it's own model size rank, its competitive performance with fewer parameters and computational resources highlights its efficiency. The use of bidirectional ViM blocks and token labeling techniques enables DeMansia Tiny to deliver robust results, making it a compelling choice for resource-constrained environments.

\section{Future Works}
Due to resource constraints, our experiments were limited to just the DeMansia Tiny variant. Looking forward, we aim to explore the performance across the entire lineup of DeMansia models with well-tuned hyperparameters. These comprehensive tests will provide deeper insights into the scalability and robustness of our approach across different scales.

Beyond expanding the experimental scope within image classification, we also note the applications of DeMansia in semantic segmentation tasks. Investigating these capabilities could open new avenues for employing DeMansia in a broader range of computer vision challenges.

Furthermore, DeMansia's adaptability as a feature extraction backend presents intriguing possibilities. We are keen on exploring how DeMansia could enhance architectures like co-detr \cite{co-detr}.

During experiments, we experience unexpected behaviour when using gradient accumulation \cite{gradient_accumulation} to simulate large batches, where there signs of extremely slow convergence. A detail analysis should be done to explore the reason behind this behaviour.

\section{Conclusions}
This research presented the development and evaluation of the DeMansia model, an adaptation and extension of the ViM \cite{vim} architecture combined with techniques from LV-ViT \cite{lv-vit} for enhanced performance in image classification tasks. The DeMansia Tiny variant demonstrates promising efficiency compared to various transformer-based architectures and shows competitive performance against more established models such as ResNet-152 \cite{resnet} and DeiT \cite{deit}.

DeMansia Tiny achieves a top-1 accuracy of 79.4\% and a top-5 accuracy of 94.5\% on the ImageNet-1k validation set, showcasing its ability to effectively handle image classification tasks with a relatively small parameter count and computational footprint. This performance positions DeMansia Tiny competitively among models of similar scale, highlighting its potential as a viable solution for resource-constrained environments.

However, it is essential to acknowledge the limitations of DeMansia Tiny. The added computational and memory overhead due to the aux head makes it more resource-intensive than some other models in its parameter range. Additionally, further fine-tuning could enhance DeMansia's performance, potentially closing the gap with larger models.

In conclusion, DeMansia offers a significant contribution to the ongoing development of efficient and effective transformer-based architectures. Its competitive performance, combined with its potential for further optimization and application across various computer vision tasks, underscores its value as a versatile and powerful model in the landscape of deep learning.

\bibliographystyle{IEEEtran}
\bibliography{references.bib}

\appendices
\section{Manual for DeMansia's Source Code}
To successfully utilize our DeMansia source code, ensure the following prerequisites are met: access to a NVIDIA GPU, and a Linux-based system equipped with the Miniforge package manager.

We provide a straightforward setup script named "setup.sh". This script automates the installation of the necessary Conda environment, integrating CUDA 11.8, Python 3.11, and the latest compatible version of Pytorch. Additionally, it compiles the custom Mamba package from \cite{vim} that supports bidirectional processing.
The repository includes a script linked in the README for setting up the ImageNet-1k dataset \cite{imagenet} after you have downloaded the required training and validation sets. The token labeling dataset, generated using \cite{nfnet} as detailed in \cite{lv-vit}, is also prepared for easy decompression.

For training, a Jupyter notebook titled "DeMansia train.ipynb" is available. It contains all the necessary code to train the DeMansia model and log various metrics such as learning rate, top-1 and top-5 accuracies on the ImageNet-1k \cite{imagenet} validation set, along with training and validation losses. For modifying the parameters that is logged, please follow the Pytorch Lightning documentation and modify the "model.py". For configuring different variants of the DeMansia model, presets are provided in "model\_config.py". You can modify these presets or add new ones according to your requirements.

For inference, load the model using the PyTorch Lightning API, which handles all hyperparameter settings automatically. Sample code for this process is provided in the README for ease of use.

\section{Training log}
The complete and interactive training log for the DeMansia Tiny model is accessible online. You can view it at \href{https://wandb.ai/catalpa/DeMansia%20Tiny}{https://wandb.ai/catalpa/DeMansia\%20Tiny}.
			
\end{document}